\documentclass[10pt,twocolumn,letterpaper]{article}

\usepackage{cvpr}
\usepackage{times}
\usepackage{epsfig}
\usepackage{graphicx}
\usepackage{amsmath}
\usepackage{amssymb}

\usepackage{pifont}
\usepackage{wrapfig}
\usepackage{boldline}
\usepackage{makecell}
\usepackage{multirow}
\usepackage{indentfirst}
\usepackage{comment}
\usepackage[font=small]{subfig}
\usepackage{color}
\usepackage{booktabs}

\newcommand\blfootnote[1]{%
  \begingroup
  \renewcommand\thefootnote{}\footnote{#1}%
  \addtocounter{footnote}{-1}%
  \endgroup
}

\usepackage[pagebackref=true,breaklinks=true,colorlinks=true,bookmarks=false]{hyperref}

\cvprfinalcopy % *** Uncomment this line for the final submission

\def\cvprPaperID{1167} % *** Enter the CVPR Paper ID here

\begin{document}

%%%%%%%%% TITLE
% \title{FlowNet3D: Learning Scene Flow in 3D Point Clouds}
\title{FlowNet3D: Learning Scene Flow in 3D Point Clouds}

\author{Xingyu Liu$^{* 1}$\qquad Charles R. Qi$^{* 2}$ \qquad Leonidas J. Guibas$^{1,2}$\\$^1$Stanford University \qquad $^2$Facebook AI Research}

\maketitle

%\thispagestyle{empty}

% %%%%%%%%% ABSTRACT
\begin{abstract}
Many applications in robotics and human-computer interaction can benefit from understanding 3D motion of points in a dynamic environment, widely noted as scene flow. While most previous methods focus on stereo and RGB-D images as input, few try to estimate scene flow directly from point clouds. In this work, we propose a novel deep neural network named FlowNet3D that learns scene flow from point clouds in an end-to-end fashion. Our network simultaneously learns deep hierarchical features of point clouds and flow embeddings that represent point motions, supported by two newly proposed learning layers for point sets. We evaluate the network on both challenging synthetic data from FlyingThings3D and real Lidar scans from KITTI. Trained on synthetic data only, our network successfully generalizes to real scans, outperforming various baselines and showing competitive results to the prior art. We also demonstrate two applications of our scene flow output (scan registration and motion segmentation) to show its potential wide use cases.

\end{abstract}

% %%%%%%%%% BODY TEXT
 \section{Introduction}
\label{sec:intro}
\blfootnote{* indicates equal contributions.}

Scene flow is the 3D motion field of points in the scene~\cite{vedula1999three}. Its projection to an image plane becomes 2D optical flow. It is a low-level understanding of a dynamic environment, without any assumed knowledge of structure or motion of the scene.
With this flexibility, scene flow can serve many higher level applications. For example, it provides motion cues for object segmentation, action recognition, camera pose estimation, or even serve as a regularization for other 3D vision problems.

\begin{figure}[t!]
\centering
\includegraphics[width=\linewidth]{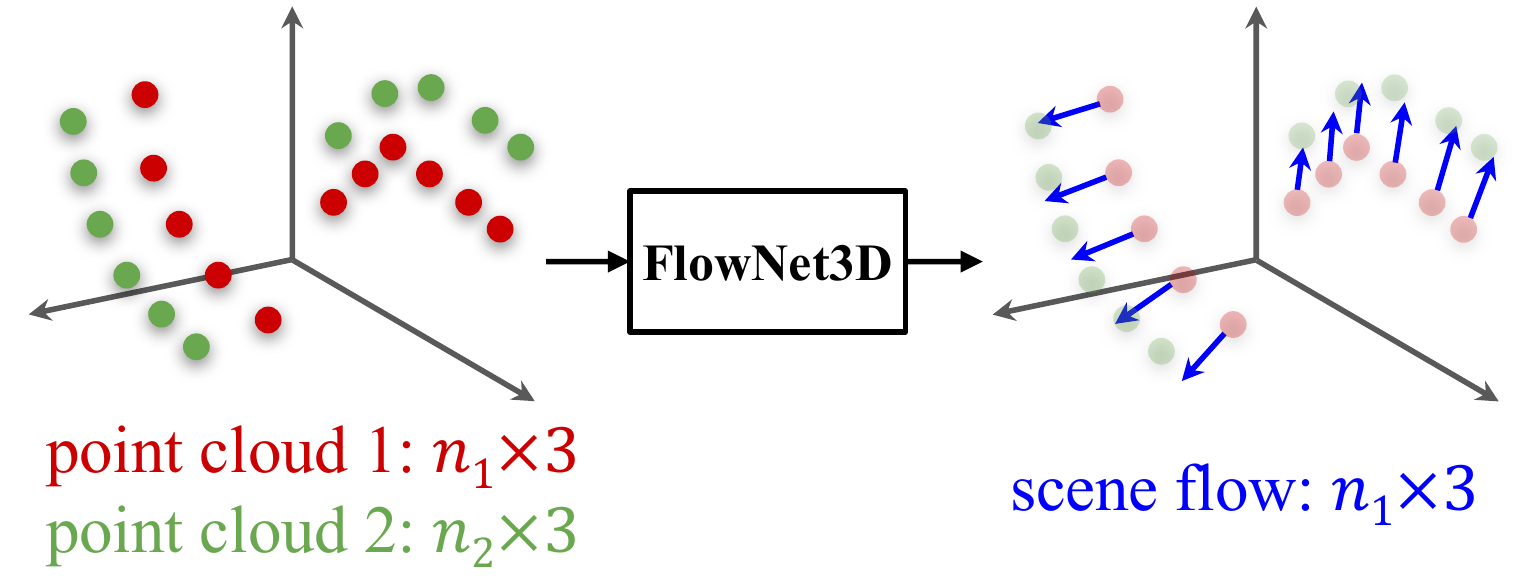}
\caption{\textbf{End-to-end scene flow estimation from point clouds.} Our model directly consumes raw point clouds from two consecutive frames, and outputs dense scene flow (as translation vectors) for all points in the 1st frame.}
\label{fig:teaser}
\end{figure}

However, for this \emph{3D} flow estimation problem, most previous works rely on \emph{2D} representations. They extend methods for optical flow estimation to stereo or RGB-D images, and usually estimate optical flow and disparity map separately~\cite{wedel2011stereoscopic,vogel20113d,OSF}, not directly optimizing for 3D scene flow. These methods cannot be applied to cases where point clouds are the only input.

Very recently, researchers in the robotics community started to study scene flow estimation directly in 3D point clouds (e.g. from Lidar)~\cite{dewan2016rigid,ushani2017learning}. But those works did not benefit from deep learning as they built multi-stage systems based on hand-crafted features, with simple models such as logistic regression. There are often many assumptions involved such as assumed scene rigidity or existence of point correspondences, which make it hard to adapt those systems to benefit from deep networks. On the other hand, in the learning domain, Qi et al.~\cite{qi2017pointnet,qi2017pointnetplusplus} recently proposed novel deep architectures that directly consume point clouds for 3D classification and segmentation. However, their work focused on processing static point clouds. 

In this work, we connect the above two research frontiers by proposing a deep neural network called \emph{FlowNet3D} that learns scene flow in 3D point clouds end-to-end.
As illustrated in Fig.~\ref{fig:teaser}, given input point clouds from two consecutive frames (point cloud 1 and point cloud 2), our network estimates a translational flow vector for every point in the first frame to indicate its motion between the two frames. 
The network, based on the building blocks from~\cite{qi2017pointnet}, is able to simultaneously learn deep hierarchical features of point clouds and \emph{flow embeddings} that represent their motions. While there are no correspondences between the two sampled point clouds, our network learns to associate points from their spatial localities and geometric similarities, through our newly proposed \emph{flow embedding} layer. Each output embedding implicitly represents the 3D motion of a point. From the embeddings, the network further up-samples and refines them in an informed way through another novel \emph{set upconv} layer. Compared to direct feature up-sampling with 3D interpolations, the set upconv layers \emph{learn} to up-sample points based on their spatial and feature relations.

We extensively study the design choices in our model and validate the usefullness of our newly proposed point set learning layers, with a large-scale synthetic dataset (FlyingThings3D). We also evaluate our model on the real LiDAR scans from the KITTI benchmark, where our model shows significantly stronger performance compared to baselines of non-deep learning methods and competitive results to the prior art. More remarkably, we show that our network, even trained on synthetic data, is able to robustly estimate scene flow in point clouds from real scans, showing its great generalizability. With fine tuning on a small set of real data, the network can achieve even better performance.

To support future research based on our work, we will release our prepared data and code for public use.

The key contributions of this paper are as follows:
\begin{itemize}
    \item We propose a novel architecture called FlowNet3D that estimates scene flow from a pair of consecutive point clouds end-to-end.
    \item We introduce two new learning layers on point clouds: a flow embedding layer that learns to correlate two point clouds, and a set upconv layer that learns to propagate features from one set of points to the other.
    \item We show how we can apply the proposed FlowNet3D architecture on real LiDAR scans from KITTI and achieve greatly improved results in 3D scene flow estimation compared with traditional methods.
\end{itemize}

 \section{Related Work}
\label{sec:related}
\textbf{Scene flow from RGB or RGB-D images.}
Vedula et al.~\cite{vedula1999three} first introduced the concept of scene flow, as three-dimensional field of motion vectors in the world. They assumed knowledge of stereo correspondences and combined optical flow and first-order approximations of depth maps to estimate scene flow. Since this seminal work, many others have tried to jointly estimate structure and motion from stereoscopic images~\cite{huguet2007variational,pons2007multi,wedel2008efficient,valgaerts2010joint,vcech2011scene,wedel2011stereoscopic,vogel20113d,vogel2013piecewise,basha2013multi,PRSM,OSF}, mostly in a variational setting with regularizations for smoothness of motion and structure~\cite{huguet2007variational,basha2013multi,valgaerts2010joint}, or with assumption of the rigidity of the local structures~\cite{vogel2013piecewise,OSF,PRSM}.

With the recent advent of commodity depth sensors, it has become feasible to estimate scene flow from monocular RGB-D images~\cite{Kinecting:the:dots}, by generalizing variational 2D flow algorithms to 3D~\cite{herbst2013rgb,jaimez2015primal} and exploiting more geometric cues provided by the depth channel~\cite{SRSF,hornacek2014sphereflow,Layered:RGBD:scene:flow}. Our work focuses on learning scene flow directly from \emph{point clouds}, without any dependence on RGB images or assumptions on rigidity and camera motions.

\noindent
\textbf{Scene flow from point clouds.}
Recently, Dewan et al.~\cite{dewan2016rigid} proposed to estimate dense rigid motion fields in 3D LiDAR scans. They formulate the problem as an energy minimization problem of a factor graph, with hand-crafted SHOT~\cite{tombari2010unique} descriptors for correspondence search.
Later, Ushani et al.~\cite{ushani2017learning} presented a different pipeline:
They train a logistic classifier to tell whether two columns of occupancy grids correspond and formulate an EM algorithm to estimate a locally rigid and non-deforming flow. Compared to these two previous works, our method is an end-to-end solution with deep learned features and no dependency on hard correspondences or assumptions on rigidity.

Concurrent to our work,~\cite{behl2018pointflownet} estimate scene flow as rigid motions of individual objects or background with network that jointly learns to regress ego-motion and detect 3D objects. \cite{shao2018motion} jointly estimate object rigid motions and segment them based on their motions. Compared to those works, our formulation does not rely on semantic supervision and focuses on solving the scene flow problem.

\noindent
\textbf{Related deep learning based methods.}
FlowNet~\cite{dosovitskiy2015flownet} and FlowNet 2.0~\cite{ilg2017flownet} are two seminal works that propose to learn optical flow with convolutional neural networks in an end-to-end fashion, showing competitive performance with great efficiency.~\cite{Flyingthings3D:Driving} extends FlowNet to simultaneously estimating disparity and optical flow. Our work is inspired by the success of those deep learning based attempts at optical flow prediction, and can be viewed as the 3D counterpart of them. However, the irregular structure in point clouds (no regular grids as in image) presents new challenges and opportunities for design of novel architectures, which is the focus of this work.

\begin{figure*}[t!]
    \centering
    \includegraphics[width=\linewidth]{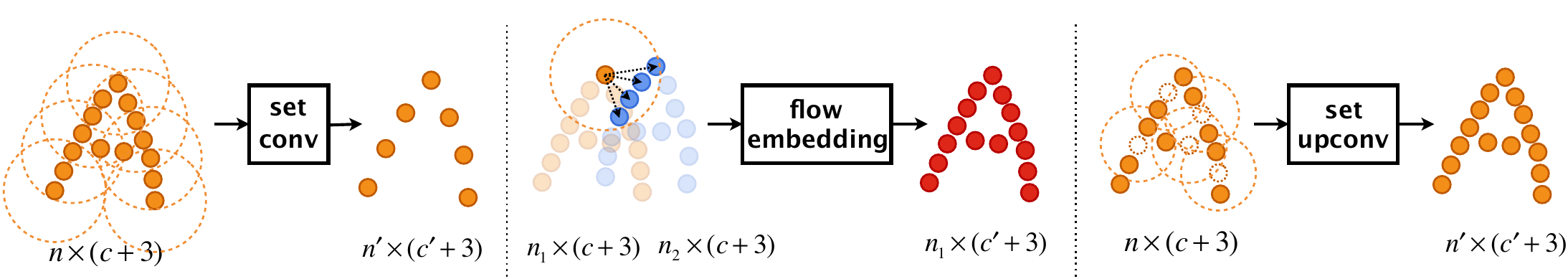}
    \caption{\textbf{Three trainable layers for point cloud processing.} \emph{Left:} the \emph{set conv} layer to learn deep point cloud features. \emph{Middle:} the \emph{flow embedding layer} to learn geometric relations between two point clouds to infer motions. \emph{Right:} the \emph{set upconv} layer to up-sample and propagate point features in a learnable way.}
    \label{fig:layers}
\end{figure*}

 \section{Problem Definition}
\label{sec:problem}

We design deep neural networks that estimate 3D motion flow from consecutive frames of point clouds. Input to our network are two sets of points sampled from a dynamic 3D scene, at two consecutive time frames: $\mathcal{P} = \{ x_i | i = 1, \ldots, n_1\}$ (point cloud 1) and $\mathcal{Q} = \{ y_j | j = 1, \ldots, n_2\}$ (point cloud 2), where $x_i, y_j \in \mathbb{R}^3$ are $XYZ$ coordinates of individual points. Note that due to object motion and viewpoint changes, the two point clouds do not necessarily have the same number of points or have any correspondences between their points. It is also possible to include more point features such as color and Lidar intensity. For simplicity we focus on $XYZ$ only.

Now consider the physical point under a sampled point $x_i$ moves to location $x'_i$ at the second frame, then the translational motion vector of the point is $d_i = x'_i - x_i$. Our goal is, given $\mathcal{P}$ and $\mathcal{Q}$, to recover the scene flow for every sampled point in the first frame: $\mathcal{D} = \{d_i | i = 1,\ldots, n_1\}$.

 \section{FlowNet3D Architecture}
\label{sec:arch}

\begin{figure*}[t!]
\centering
\includegraphics[width=\linewidth]{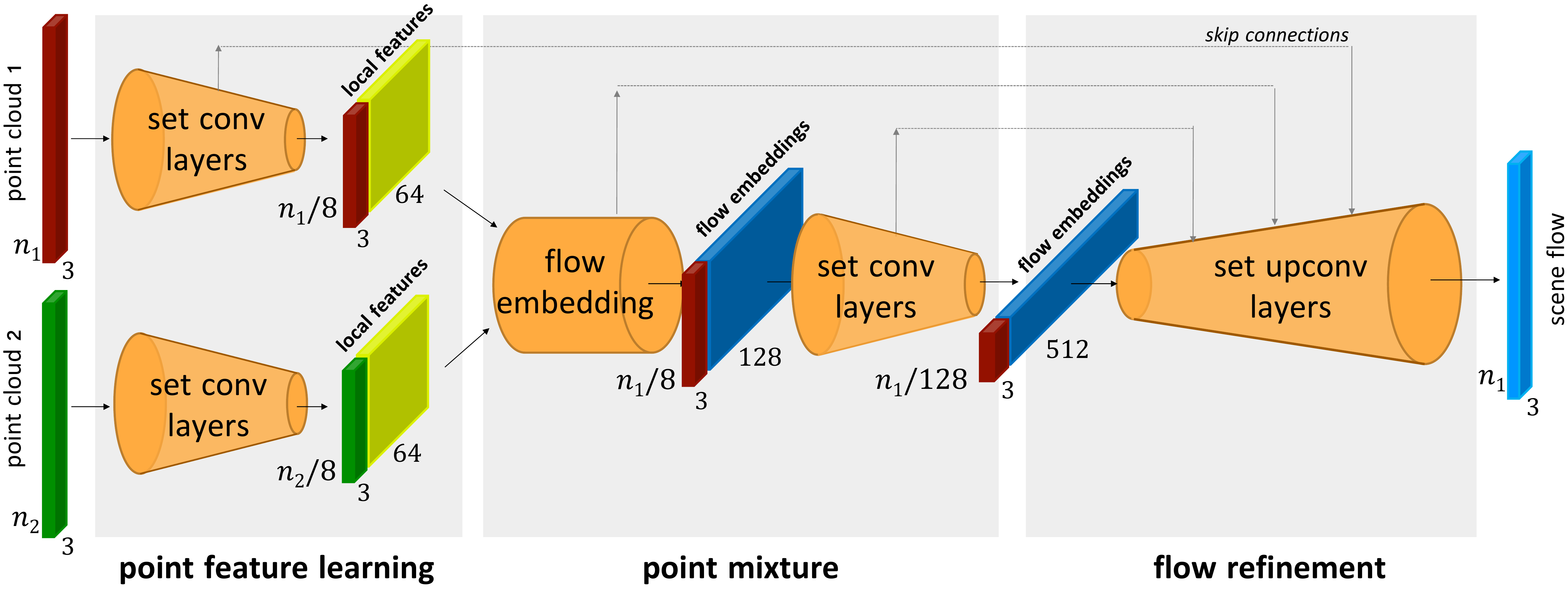}
\caption{\textbf{FlowNet3D architecture.} Given two frames of point clouds, the network learns to predict the scene flow as translational motion vectors for each point of the first frame. See Fig.~\ref{fig:layers} for illustrations of the layers and Sec.~\ref{sec:flownet3d:arch} for more details on the network architecture.}
\label{fig:network_architecture}
\end{figure*}

In this section, we introduce FlowNet3D (Fig.~\ref{fig:network_architecture}), an end-to-end scene flow estimation network on point clouds. The model has three key modules for (1) point feature learning, (2) point mixture, and (3) flow refinement. Under these modules are three key deep point cloud processing layers: \emph{set conv} layer, \emph{flow embedding} layer and \emph{set upconv} layer (Fig.~\ref{fig:layers}). In the following subsections, we describe each modules with their associating layers in details, and specify the final FlowNet3D architecture in Sec.~\ref{sec:flownet3d:arch}.

\subsection{Hierarchical Point Cloud Feature Learning}
\label{sec:flownet3d:feature}
Since a point cloud is a \emph{set} of points that is irregular and orderless, traditional convolutions do not fit.
We therefore follow a recently proposed PointNet++ architecture~\cite{qi2017pointnetplusplus}, a translation-invariant network that learns hierarchical features. Although the set conv layer~\footnote{Noted as set abstraction layer in~\cite{qi2017pointnetplusplus}. We name it set conv here to emphasize its spatial locality and translation invariance.} was designed for 3D classification and segmentation, we find its feature learning layers also powerful for the task of scene flow.

As shown in Fig.~\ref{fig:layers} (left), a set conv layer takes a point cloud with $n$ points, each point $p_i = \{x_i, f_i\}$ with its $XYZ$ coordinates $x_i \in \mathbb{R}^3$ and its feature $f_i \in \mathbb{R}^c$ ($i=1,...,n$), and outputs a sub-sampled point cloud with $n'$ points, where each point $p'_j = \{x'_j, f'_j\}$ has its $XYZ$ coordinates $x'_j$ and an updated point feature $f_j' \in \mathbb{R}^{c'}$ ($j=1,...n'$).

Specifically, as described more closely in~\cite{qi2017pointnetplusplus}, the layer firstly samples $n'$ regions from the input points with farthest point sampling (with region centers as $x'_j$), then for each region (defined by a radius neighborhood specified by radius $r$), it extracts its local feature with the following symmetric function
\begin{equation}
    f'_j = \underset{\{i | \lVert x_i - x'_j \lVert \leq r\}}{\mbox{MAX}}\left\{h(f_i, x_i - x'_j)\right\}.
    \label{eq:setconv}
\end{equation}
where $h: \mathbb{R}^{c+3} \rightarrow \mathbb{R}^{c'}$ is a non-linear function (realized as a multi-layer perceptron) with concatenated $f_i$ and $x_i-x'_j$ as inputs, and $\mbox{MAX}$ is element-wise max pooling.

\subsection{Point Mixture with Flow Embedding Layer}
\label{sec:flownet3d:mixture}

To mix two point clouds we rely on a new \emph{flow embedding} layer (Fig.~\ref{fig:layers} middle). To inspire our design, imagine a point at frame $t$, if we know its \emph{corresponding} point in frame $t+1$ then its scene flow is simply their relative displacement. However, in real data, there are often no correspondences between point clouds in two frames, due to viewpoint shift and occlusions. It is still possible to estimate the scene flow though, because we can find multiple softly corresponding points in frame $t+1$ and make a ``weighted'' decision.

 Our \emph{flow embedding} layer learns to aggregate both (geometric) feature similarities and spatial relationships of points to produce embeddings that encode point motions. Compared to the set conv layer that takes in a single point cloud, the flow embedding layer takes a \emph{pair} of point clouds: $\{p_i = (x_i, f_i)\}_{i=1}^{n_1}$ and $\{q_j = (y_j, g_j)\}_{j=1}^{n_2}$ where each point has its $XYZ$ coordinate $x_i, y_j \in \mathbb{R}^3$, and a feature vector $f_i, g_j \in \mathbb{R}^c$. The layer learns a flow embedding for each point in the first frame: $\{ e_i\}_{i=1}^{n_1}$ where $e_i \in \mathbb{R}^{c'}$. We also pass the original coordinates $x_i$ of the points in the first frame to the output, thus the final layer output is $\{o_i = (x_i, e_i)\}_{i=1}^{n_1}$.
 
The underneath operation to compute $e_i$ is similar to the one in set conv layers. However, their physical meanings are vastly different. For a given point $p_i$ in the first frame, the layer firstly finds all the points $q_j$ from the second frame in its radius neighborhood (highlighted blue points).
 If a particular point $q^* = \{y^*, g^*\}$ corresponded to $p_i$, then the flow of $p_i$ were simply $y^* - x_i$. Since such case rarely exists, we instead use a neural layer to aggregate flow votes from all the neighboring $q_j$'s
 
 \begin{equation}
     e_i = \underset{\{j | \lVert y_j - x_i \lVert \leq r\}}{\mbox{MAX}}\left\{h(f_i, g_j, y_j - x_i)\right\}.
     \label{eq:flowembedding}
 \end{equation}
 
 \noindent
 where $h$ is a non-linear function with trainable parameters similar to the set conv layer and $\mbox{MAX}$ is the element-wise max pooling. Compared to Eq.~\eqref{eq:setconv}, we input two point features to $h$, expecting it to learn to compute the ``weights'' to aggregate all potential flow vectors $d_{ij} = y_j - x_i$.
 
 An alternative formulation is to explicitly specify how we relate point features, by computing a feature distance $dist(f_i, g_j)$. The feature distance is then fed to the non-linear function $h$ (instead directly feeding the $f_i$ and $g_j$). In ablation studies we show that our formulation in Eq.~\eqref{eq:flowembedding} learns more effective flow embeddings than this alternative.
 
The computed flow embeddings are further mixed through a few more set conv layers 
 so that we obtain spatial smoothness. This also help resolve ambiguous cases (e.g. points on the surface of a translating table) that require large receptive fields for flow estimation.

\begin{comment}
\subsubsection{Flow refinement with trainable set upconv layers.}
In this part of the network, we want to propagate flow embeddings from the subsampled points to all original points in frame 1. This is a similar task to what faced by the feature propagation stage described in~\cite{qi2017pointnetplusplus}, where features need to be upsampled for object part segmentation and scene semantic segmentation tasks. However, in the previous work~\cite{qi2017pointnetplusplus}, the feature propagation layer is not trainable. It is just a predefined 3D interpolation layer with inverse distance weights -- it has no ability in learning how to up propagate flow embeddings.

In this work, we propose a novel trainable layer for feature up propagation -- \emph{set upconv layer} as shown in Fig.~\ref{fig:layers} (right). It is very similar to set conv layer except that instead of having a farthest point sampler to subsample the points, a set upconv layer directly takes $m \times 3$ coordinates to indicate where we want to upsample to. Those $m \times 3$ coordinates are exactly those from the corresponding set conv layers in flow mixture or point feature learning parts.

Now compared to using 3D interpolation, our network learns how to weight the nearby points' features, just as how the flow embedding layer weight displacements. We will show that the new set upconv layer shows significant advantage in empirical results.
\end{comment}

\subsection{Flow Refinement with Set Upconv Layer}
\label{sec:flownet3d:refine}

In this module, we up-sample the flow embeddings associated with the intermediate points to the original points, and at the last layer predict flow for all the original points. The up-sampling step is achieved by a learnable new layer -- the \emph{set upconv} layer, which learns to propagate and refine the embeddings in an informed way.

Fig.~\ref{fig:layers} (right) illustrates the process of a set upconv layer. The inputs to the layer are source points $\{p_i = \{x_i, f_i\} |i = 1,\ldots,n\}$, and a set of target point coordinates $\{x'_j | j=1,\ldots,n'\}$ which are locations we want to propagate the source point features to. For each target location $x'_j$ the layer outputs its point feature $f'_j \in \mathbb{R}^{c'}$ (propagated flow embedding in our case) by aggregating its neighboring source points' features.

Interestingly, just like in 2D convolutions in images where upconv2D can be implemented through conv2D, our \emph{set upconv} can also be directly achieved with the same \emph{set conv} layer as defined in Eq.~\eqref{eq:setconv}, but with a different local region sampling strategy. Instead of using farthest point sampling to find $x'_j$ as in the set conv layer, we compute features on \emph{specified} locations by the target points $\{x'_j\}_{j=1}^{n'}$.

Note that although $n' > n$ in our up-sampling case, the \emph{set upconv} layer itself is flexible to take any number of target locations which unnecessarily correspond to any real points. It is a flexible and trainable layer to propagate/summarize features from one point cloud to another.

Compared to an alternative way to up-sample point features -- using 3D interpolation ($f'_j = \sum_{\{i | \lVert x_i - x'_j \lVert \leq r\}} w(x_i, x'_j) f_i$ with $w$ as a normalized inverse-distance weight function~\cite{qi2017pointnetplusplus}), our network learns how to weight the nearby points' features, just as how the flow embedding layer weights displacements. We find that the new set upconv layer shows significant advantages in empirical results.

\subsection{Network Architecture}
\label{sec:flownet3d:arch}

The final FlowNet3D architecture is composed of four set conv layers, one flow embedding layer and four set upconv layers (corresponding to the four set conv layers) and a final linear flow regression layer that outputs the $\mathbb{R}^3$ predicted scene flow. For the set upconv layers we also have skip connections to concatenate set conv output features. Each learnable layer adopts multi-layer perceptrons for the function $h$ with a few Linear-BatchNorm-ReLU layers parameterized by its linear layer width. The detailed layer parameters are as shown in Table~\ref{tab:flownet3d_spec}.

\begin{table}[]
\small
\centering
\begin{tabular}{c|c|c|c}
\hline
Layer type     & $r$   & Sample rate & MLP width   \\ \hline
set conv        & $0.5$ & $0.5\times$        & $[32,32,64]$    \\ 
set conv        & $1.0$ & $0.25\times$       & $[64,64,128]$   \\ 
flow embedding & $5.0$ & $1\times$          & $[128,128,128]$ \\ 
set conv        & $2.0$ & $0.25\times$       & $[128,128,256]$ \\ 
set conv        & $4.0$ & $0.25\times$       & $[256,256,512]$ \\ 
set upconv      & $4.0$ & $4\times$          & $[128,128,256]$ \\ 
set upconv      & $2.0$ & $4\times$          & $[128,128,256]$ \\ 
set upconv      & $1.0$ & $4\times$          & $[128,128,128]$ \\ 
set upconv      & $0.5$ & $2\times$          & $[128,128,128]$ \\ 
linear         & -   & -          & $3^*$ \\ \hline
\end{tabular}
\caption{\textbf{FlowNet3D architecture specs.} Note that the last layer is linear thus has no ReLU and batch normalization.}
\label{tab:flownet3d_spec}
\end{table}

 \section{Training and Inference wtih FlowNet3D}
\label{sec:training}

We take a supervised approach to train the FlowNet3D model with ground truth scene flow supervision. While this dense supervision is hard to acquire in real data, we tap large-scale synthetic dataset (FlyingThings3D) and show that our model trained on synthetic data generalizes well to real Lidar scans (Sec.~\ref{sec:exp:kitti}). 

\paragraph{Training loss with cycle-consistency regularization.} We use smooth $L_1$ loss (huber loss) for scene flow supervision, together with a cycle-consistency regularization. Given a point cloud $\mathcal{P} = \{x_i\}_{i=1}^{n_1}$ at frame $t$ and a point cloud $\mathcal{Q} = \{y_j\}_{j=1}^{n_2}$ at frame $t+1$, the network predicts scene flow as $\mathcal{D} = F(\mathcal{P}, \mathcal{Q}; \Theta) = \{d_i\}_{i=1}^{n_1}$ where $F$ is the FlowNet3D model with parameters $\Theta$. With ground truth scene flow $\mathcal{D^*} = \{d^*_i\}_{i=1}^{n_1}$, our loss is defined as in Eq.~\eqref{eq:loss}. In the equation, $\lVert d'_i + d_i \lVert$ is the \emph{cycle-consistency} term that enforces the \emph{backward flow} $\{d'_i\}_{i=1}^{n_1} = F(\mathcal{P'}, \mathcal{P}; \Theta)$ from the shifted point cloud $\mathcal{P'}=\{x_i+d_i\}_{i=1}^{n_1}$ to the original point cloud $\mathcal{P}$ is close to the reverse of the \emph{forward flow}
\begin{equation}
L(\mathcal{P}, \mathcal{Q}, \mathcal{D^*}, \Theta) = \frac{1}{n_1}\sum_{i=1}^{n_1} \Big\{ \lVert d_i - d^*_i \lVert + \lambda \lVert d'_i + d_i \lVert \Big\}
\label{eq:loss}
\end{equation}
\paragraph{Inference with random re-sampling.} A special challenge with regression problems (such as scene flow) in point clouds is that down-sampling introduces noise in prediction.
A simple but effective way to reduce the noise is to randomly re-sample the point clouds for multiple inference runs and average the predicted flow vectors for each point. In the experiments, we will see that this re-sampling and averaging step leads to a slight performance gain.

 \section{Experiments}
\label{sec:exp}
In this section, we first evaluate and validate our design choices in Sec.~\ref{sec:exp:flyingthings3d} with a large-scale synthetic dataset (FlyingThings3D), and then in Sec.~\ref{sec:exp:kitti} we show how our model trained on synthetic data can generalize successfully to real Lidar scans from KITTI. Finally, in Sec.~\ref{sec:exp:app} we demonstrate two applications of scene flow on 3D shape registration and motion segmentation.

\begin{table}[t!]
\small
\setlength{\tabcolsep}{4.8pt}
\centering
\begin{tabular}{c|c|ccc}
\toprule
Method   & Input & EPE & \begin{tabular}[c]{@{}c@{}}ACC \\ (0.05)\end{tabular}  & \begin{tabular}[c]{@{}c@{}}ACC \\ (0.1)\end{tabular} \\ \midrule
\multirow{2}{*}{FlowNet-C \cite{dosovitskiy2015flownet}} & depth & 0.7887  & 0.20\%  & 1.49\%  \\
 & RGBD & 0.7836 & 0.25\% & 1.74\%  \\ \midrule
ICP \cite{ICP}  & points & 0.5019 &     7.62\%  &  21.98\%    \\
EM-baseline (ours) &    points&  0.5807&  2.64\%  &  12.21\%   \\
LM-baseline (ours) &    points &  0.7876  &  0.27\%   &  1.83\% \\
DM-baseline (ours) &   points&     0.3401  &  4.87\%   &  21.01\%  \\
\midrule
FlowNet3D (ours) &    points&   \textbf{0.1694} &   \textbf{25.37\%}   & \textbf{57.85\%}   \\ 
\bottomrule
\end{tabular}
\caption{\textbf{Flow estimation results on the FlyingThings3D dataset.} Metrics are End-point-error (EPE), Acc ($<$0.05 or 5\%, $<$0.1 or 10\%) for scene flow.
}
\label{tab:flyingthing3d}
\end{table}

\subsection{Evaluation and Design Validation on FlyingThings3D}
\label{sec:exp:flyingthings3d}
As annotating or acquiring dense scene flow is very expensive on real data, there does not exist any large-scale real dataset with scene flow annotations to the best of our knowledge~\footnote{The KITTI dataset we test on in Sec.~\ref{sec:exp:kitti} only has 200 frames with annotations.~\cite{wang2018deep} mentioned a larger dataset however it belongs to Uber and is not publicly available.}. Therefore, we turn to a synthetic, yet challenging and large-scale dataset, FlyingThings3D, to train and evaluate our model as well as to validate our design choices.

\paragraph{FlyingThings3D~\cite{Flyingthings3D:Driving}.} The dataset consists of stereo and RGB-D images rendered from scenes with multiple randomly moving objects sampled from ShapeNet \cite{ShapeNet}. There are in total around 32k stereo images with ground truth disparity and optical flow maps. We randomly sub-sampled 20,000 of them as our training set and 2,000 as our test set. Instead of using RGB images, we preprocess the data by popping up disparity maps to 3D point clouds and optical flow to scene flow. We will release our prepared data.

\paragraph{Evaluation Metrics.} We use 3D end point error (EPE) and flow estimation accuracy (ACC) as our metrics. The 3D EPE measures the average $L_2$ distance between the estimated flow vector to the ground truth flow vector. Flow estimation accuracy measures the portion of estimated flow vectors that are below a specified end point error, among all the points. We report two ACC metrics with different thresholds.

\paragraph{Results.}
Table~\ref{tab:flyingthing3d} reports flow evaluation results on the test set, comparing FlowNet3D to various baselines.
Among the baselines, FlowNet-C is a CNN model adapted from \cite{ilg2017flownet} that learns to predict scene flow from a pair of depth images or RGB-D images (depth images transformed to $XYZ$ coordinate maps for input), instead of optical flow from RGB images as originally in~\cite{ilg2017flownet} (more architecture details in supplementary). 
However, we see that this image-based method has a hard time predicting accurate scene flow probably because of strong occlusions and clutters in the 2D projected views. We also compare with an ICP (iterative closest point) baseline that finds a single rigid transform for the entire scene, which matches large objects in the scene but is unable to adapt to the multiple independently moving objects in our input. Surprisingly, this ICP baseline is still able to get some reasonable numbers (even better than the 2D FlowNet-C one).

\begin{figure}[t!]
\centering
\includegraphics[width=0.9\linewidth]{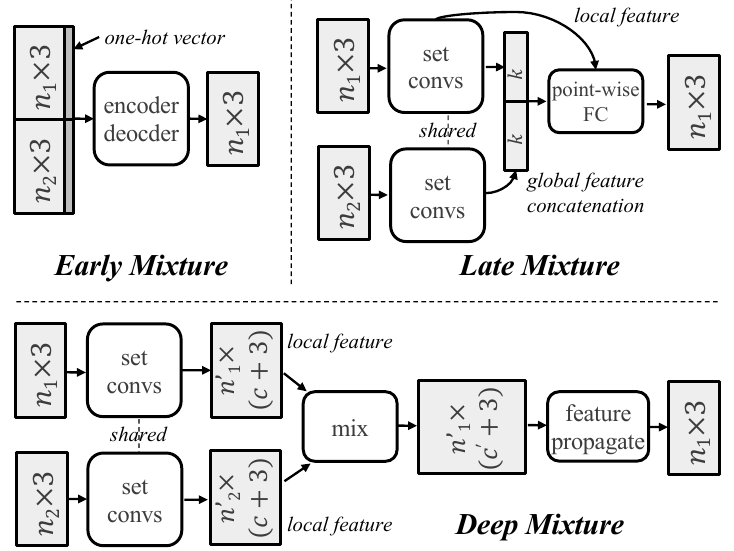}
\caption{\textbf{Three meta-architectures for scene flow network.}
FlowNet3D (Fig.~\ref{fig:network_architecture}) belongs to the deep mixture.
}
\label{fig:meta_arch}
\end{figure}

We also report results of three baseline deep models that directly consume point clouds (as instantiations of the three meta-architectures in Fig.~\ref{fig:meta_arch}). They mix point clouds of two frames at early, late, or intermediate stages. The EM-baseline combines two point clouds into a single set at input and distinguishes them by appending each point with a one-hot vector of length 2. The LM-baseline firstly computes a global feature for the point cloud from each frame, and then concatenates the global features as a way to mix the points. The DM-baseline is similar in structure to our FlowNet3D (they both belong to the DM meta-architecture) but uses a more naive way to mix two intermediate point clouds (by concatenating all features and point displacements and processing it with fully connected layers), and it uses 3D interpolation instead of set upconv layers to propagate point features.
More details are provided in the supplementary.

Compared to those baseline models, our FlowNet3D achieves much lower EPE as well as significantly higher accuracy.

\begin{table}[t!]
\small
\centering
\setlength{\tabcolsep}{4pt}
\begin{tabular}{c|c|c|c|c|c}
\toprule
\multirow{2}{*}{\begin{tabular}[c]{@{}c@{}}Feature \\ distance \end{tabular}} & \multirow{2}{*}{Pooling} & \multirow{2}{*}{Refine} & \multirow{2}{*}{\begin{tabular}[c]{@{}c@{}} Multiple \\ resample\end{tabular}} & \multirow{2}{*}{\begin{tabular}[c]{@{}c@{}} Cycle- \\ consistency \end{tabular}} & \multirow{2}{*}{EPE} \\ 
&&&& &\\ \midrule
dot &  avg &  interp &    {\color{black} \ding{55}} & {\color{black} \ding{55}} &    0.3163    \\ \midrule 
dot  &   max&    interp &   {\color{black} \ding{55}} & {\color{black} \ding{55}} &  0.2463    \\ 
cosine   &   max &  interp &    {\color{black} \ding{55}} & {\color{black} \ding{55}} &  0.2600   \\
learned   &   max &   interp  &  {\color{black} \ding{55}} & {\color{black} \ding{55}} &  0.2298    \\ \midrule 
learned& max & upconv &   {\color{black} \ding{55}} & {\color{black} \ding{55}} &   0.1835 \\ 
learned & max & upconv &  {\color{black} \ding{51}} & {\color{black} \ding{55}} &  0.1694 \\
learned & max & upconv &  {\color{black} \ding{51}} & {\color{black} \ding{51}} &  \textbf{0.1626} \\
\bottomrule
\end{tabular}
\caption{\textbf{Ablation studies on the FlyingThings3D dataset.} We study the effects of distance function, type of pooling in $h$, layers used in flow refinement, as well as effects of re-sampling and cycle-consistency regularization.}
\label{tab:ablation}
\end{table}

\paragraph{Ablation studies.}
Table~\ref{tab:ablation} shows the effects of several design choices of FlowNet3D.
Comparing the first two rows, we see max pooling has a significant advantage over average pooling, probably because max pooling is more selective in picking ``corresponding'' point and suffers less from noise. From row 2 to row 4, we compare our design to the alternatives of using feature distance functions (as discussed in Sec.~\ref{sec:flownet3d:mixture}) with cosine distance and its unnormalized version (dot product).
Our approach gets the best performance, with (\emph{11.6\%} error reduction compared to using the cosine distance. Looking at row 4 and row 5, we see that our newly proposed set upconv layer significantly reduces flow error by \emph{20\%}. Lastly, we find multiple re-sampling (10 times) during inference (second last row) and training with cycle-consistency regularization (with $\lambda = 0.3$) further boost the performance. The final row represents the final setup of our FlowNet3D.

\begin{table}[t!]
\small
\setlength{\tabcolsep}{1.6pt}
\centering
\begin{tabular}{c | c | c | c | c }
\toprule
Method   & Input  & \begin{tabular}[c]{@{}c@{}} EPE\\ (meters)\end{tabular}  & \begin{tabular}[c]{@{}c@{}} outliers\\ (0.3m or 5\%)\end{tabular} & \begin{tabular}[c]{@{}c@{}}KITTI \\ ranking \end{tabular}   \\ \midrule 
LDOF \cite{LDOF} & RGB-D & 0.498 & 12.61\% & 21 \\
OSF \cite{OSF} &  RGB-D  & 0.394  & 8.25\% & 9 \\ 
\multirow{2}{*}{PRSM \cite{PRSM}} &  RGB-D &  0.327  & 6.06\% & \multirow{2}{*}{3} \\
 &  RGB stereo & 0.729  & 6.40\% & \\ 
\midrule 
Dewan et al. \cite{dewan2016rigid} & points & 0.587 & 71.74\% & - \\ 
ICP (global) & points & 0.385 & 42.38\% & - \\ 
ICP (segmentation) &  points & 0.215 & 13.38\% & - \\ 
\midrule 
FlowNet3D (ours) &  points & \textbf{0.122}  &  \textbf{5.61\%} & - \\ 
\bottomrule 
\end{tabular}
\caption{\textbf{Scene flow estimation on the KITTI scene flow dataset (w/o ground points).}
Metrics are EPE, outlier ratio ($>$0.3m or 5\%). KITTI rankings are the methods' rankings on the KITTI scene flow leaderboard. 
Our FlowNet3D model is trained on the synthetic FlyingThings3D dataset.
}
\label{tab:kitti_results}
\end{table}

\begin{figure*}[t!]
    \centering    
\includegraphics[width=0.9\textwidth]{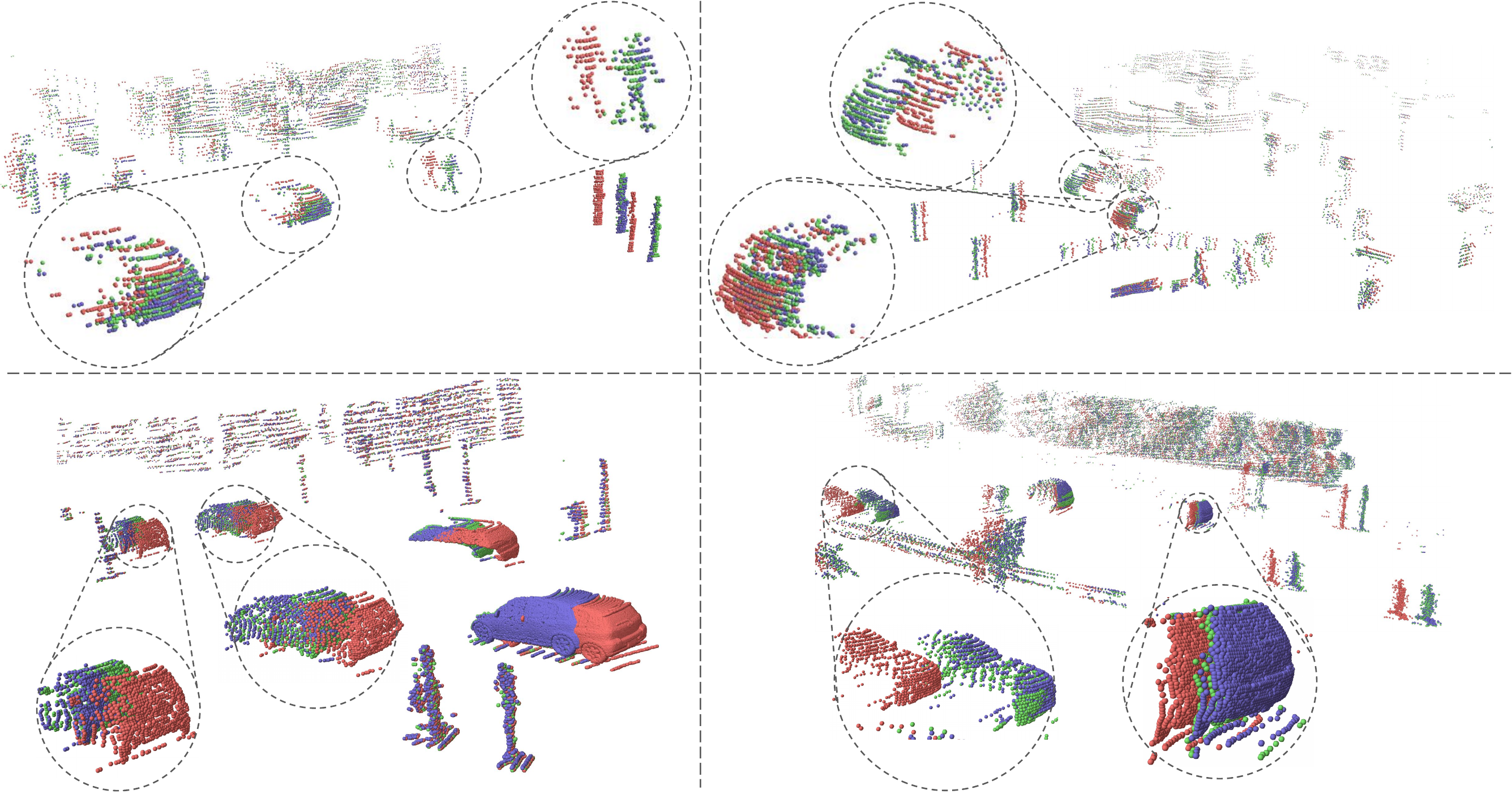}
\caption{\textbf{Scene flow on KITTI point clouds.} We show scene flow predicted by FlowNet3D on four KITTI scans. Lidar points are colored to indicate points as from \textcolor{red}{frame 1}, \textcolor{green}{frame 2} or as \textcolor{blue}{translated points} (point cloud 1 + scene flow).}
\label{fig:kitti}
\end{figure*}

\begin{table*}[t!]
\small
\centering
\begin{tabular}{c|c|c|c|c|c|c}
\toprule
\multirow{2}{*}{Method} & \multirow{2}{*}{\begin{tabular}[c]{@{}c@{}} PRSM \cite{PRSM} \\ (RGB stereo) \end{tabular}} & \multirow{2}{*}{\begin{tabular}[c]{@{}c@{}} PRSM \cite{PRSM} \\ (RGB-D) \end{tabular}} &\multirow{2}{*}{\begin{tabular}[c]{@{}c@{}} ICP \\ (global) \end{tabular}} &
\multirow{2}{*}{\begin{tabular}[c]{@{}c@{}} FlowNet3D \\ (without finetune) \end{tabular}} & \multirow{2}{*}{\begin{tabular}[c]{@{}c@{}} FlowNet3D + ICP \\ (without finetune) \end{tabular}} &     
\multirow{2}{*}{\begin{tabular}[c]{@{}c@{}} FlowNet3D \\ (with finetune) \end{tabular}} \\
&&&&\\ \midrule
3D EPE & 0.668 & 0.368 & 0.281 & 0.211 & 0.195 & \textbf{0.144}  \\ \midrule
3D outliers & 6.42\% & \textbf{6.06\%} & 24.29\% & 20.71\% & 13.41\% & 9.52\% \\
\bottomrule
\end{tabular}
\caption{\textbf{Scene flow estimation on the KITTI sceneflow dataset (w/ ground points).} The first 100 frames are used to finetune our model. All methods are evaluated on the rest 50 frames.}

\label{tab:kitti_with_ground}
\end{table*}

\subsection{Generalization to Real Lidar Scans in KITTI}
\label{sec:exp:kitti}

In this section, we show that our model, trained on the synthetic dataset, can be directly applied to detect scene flow in point clouds from real Lidar scans from KITTI.

\paragraph{Data and setup.} We use the KITTI scene flow dataset~\cite{Menze2015ISA,OSF}, which is designed for evaluations of RGB stereo based methods. To evaluate point cloud based method, we use its ground truth labels and trace raw point clouds associated to the frames. Since no point cloud is provided for the test set (and part of the train set), we evaluate on all 150 out of 200 frames from the \emph{train set} with available point clouds.
Furthermore, to keep comparison fair with the previous method~\cite{dewan2016rigid}, we firstly evaluation our model on Lidar scans with removed grounds~\footnote{The ground is a large piece of flat geometry that provides little cue to its motion but at the same time occupies a large portion of points, which biases the evaluation results.} (see supplementary for details) in Table~\ref{tab:kitti_results}. We then report another set of results with the full Lidar scans including the ground points in Table~\ref{tab:kitti_with_ground}.

\paragraph{Baselines.} LDOF+depth~\cite{LDOF} uses a variational model to solve optical flow and treats depth as an extra feature dimension. OSF~\cite{OSF} uses discrete-continuous CRF on superpixels with the assumption of rigid motion of objects. PRSM~\cite{PRSM} uses energy minimization on rigidly moving segments and jointly estimates multiple attributes together including rigid motion. Since the three RGB-D image based methods do not output scene flow directly (but optical flow and disparity separately), we either use estimated disparity (fourth row) or pixel depth change (first three rows) to compute depth-wise flow displacements.

ICP (global) estimates a single rigid motion for the entire scene. ICP (segmentation) is a stronger baseline that first computes connected components on Lidar points after ground removal and then estimates rigid motions for each individual segment of point clouds.

\paragraph{Results.} In Table~\ref{tab:kitti_results}, we compare FlowNet3D with prior arts optimized for 2D optical flow as well as the two ICP baselines on point clouds.
Compared to 2D-image based methods \cite{LDOF,OSF,PRSM}, our method shows great advantages on scene flow estimation -- achieving significantly lower 3D end-point error (\emph{63\% relative error reduction} from~\cite{PRSM}) and 3D outlier ratios.
Our method also outperforms the two ICP baselines that rely more on rigidity of global scene or correctness of segmentation.
Additionally, we conclude that our model, although only trained on synthetic data, remarkably generalizes well to the real Lidar point clouds.

Fig.~\ref{fig:kitti} visualizes our scene flow prediction. We can see our model can accurately estimate flows for dynamic objects, such as moving vehicles and pedestrians.

In Table~\ref{tab:kitti_with_ground} we report results on the full Lidar scans with ground point clouds. We also split the data to use 100 frames to finetune our FlowNet3D model on Lidar scans, and use the rest 50 for testing. We see that including ground points negatively impacted all methods. But our method still outperforms the ICP baseline. By adopting ICP estimated flow on the segmented grounds and net estimated flow for the rest of points (FlowNet3D+ICP), our method can also beat the prior art (PRSM) in EPE. The PRSM leads in outlier ratio because flow estimation for grounds is more friendly with methods taking images input. By finetuning FlowNet3D on the Lidar scans, our model even achieves better results (the last column).

\begin{figure}[t!]
    \centering
    \includegraphics[width=\linewidth]{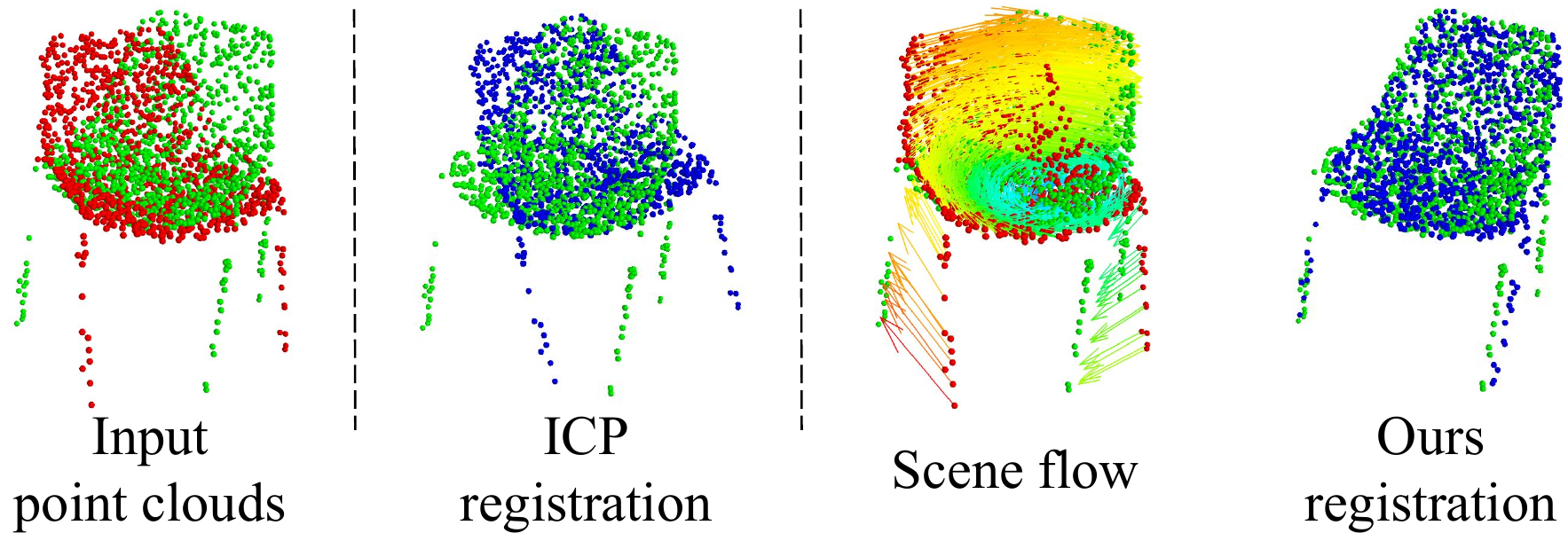}
    \caption{\textbf{Partial scan registration of two chair scans.} The goal is to register point cloud 1 (red) to point cloud 2 (green). The transformed point cloud 1 is in blue. We show a case where ICP fails to align the chair while our method grounded by dense scene flow succeeds.}
    \label{fig:registration}
\end{figure}

\begin{table}[t!]
\centering
\begin{tabular}{c|c|c|c}
\toprule
    & ICP   &  Scene flow (SF)  & SF + Rigid motion \\ \midrule
EPE & 0.384 & 0.220 & 0.125    \\ \bottomrule
\end{tabular}
\caption{Point cloud warping errors.}
\label{tab:registration}
\end{table}

\subsection{Applications}
\label{sec:exp:app}

While scene flow itself is a low-level signal in understanding motions, it can provide useful cues for many higher level applications as shown below (more details on the demo and datasets are included in supplementary).

\subsubsection{3D Scan Registration}

Point cloud registration algorithms (e.g. ICP) often rely on finding correspondences between the two point sets. However due to scan partiality, there are often no direct correspondences. In this demo, we explore in using the dense scene flow predicted from FlowNet3D for scan registration. The point cloud 1 shifted by our predicted scene flow has a natural correspondence to the original point cloud 1 and thus can be used to estimate a rigid motion between them. We show in Fig.~\ref{fig:registration} that in partial scans our scene flow based registration can be more robust than the ICP method in cases when ICP stucks at a local minimum. Table~\ref{tab:registration} quantitatively compares the 3D warping error (the EPE from warped points to ground truth points) of ICP, directly using our scene flow and using scene flow followed by a rigid motion estimation.

\subsubsection{Motion Segmentation}

\begin{figure}[t!]
    \centering
    \includegraphics[width=\linewidth]{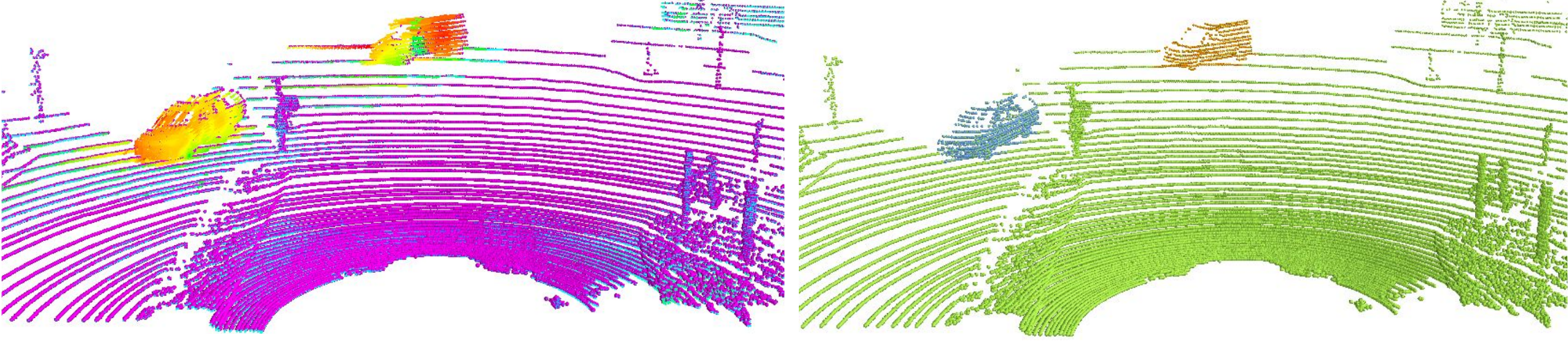}
    \caption{\textbf{Motion segmentation of a Lidar point cloud.} \emph{Left:} Lidar points and estimated scene flow in colored quiver vectors. \emph{Right:} motion segmented objects and regions.}
    \label{fig:motion_seg}
\end{figure}

Our estimated scene flow in Lidar point clouds can also be used for motion segmentation of the scene -- segmenting the scene into different objects or regions depending on their motions. In Fig.~\ref{fig:motion_seg}, we demonstrate motion segmentation results in a KITTI scene, where we clustered Lidar points based on their coordinates and estimated scene flow vectors. We see that different moving cars, grounds, and static objects are clearly segmented from each other. Recently,~\cite{shao2018motion} also tried to jointly estimate scene flow and motion segmentation from RGB-D input. It is interesting to augment our pipeline for similar tasks in point clouds in the future.

 \section{Conclusion}
\label{sec:conclusion}
In this paper, we have presented a novel deep neural network architecture that estimates scene flow directly from 3D point clouds, as arguablely the first work that shows success in solving the problem end-to-end with point clouds. To support FlowNet3D, we have proposed a novel flow embedding layer that learns to aggregate geometric similarities and spatial relations of points for motion encoding, as well as a new set upconv layer for trainable set feature propagation. On both challenging synthetic dataset and real Lidar point clouds, we validated our network design and showed its competitive or better results to various baselines and prior arts. We have also demonstrated two example applications of using scene flow estimated from our model.

\section*{Acknowledgement}

This research was supported by Toyota-Stanford AI Center grant TRI-00387, NSF grant IIS-1763268, a Vannevar Bush Faculty Fellowship, and a gift from Amazon AWS.

%%%%%%%%% BIB
{\small
\bibliographystyle{ieee}
\bibliography{egbib}
}

\newpage

%%%%%%%%% SUPPLEMENTARY TEXT
\appendix
\section*{Supplementary}

\section{Overview}
In this document, we provide more details to the main paper and show extra results on model size, running time and feature visualization.

In Sec.~\ref{sec:flying} we describe details in the FlyingThings3D experiments. In Sec.~\ref{sec:arch_detail}, we provide more details on the baseline architectures (main paper Sec. 6.1). In Sec.~\ref{sec:kitti_prep} we describe how we prepared KITTI Lidar scans for our evaluations (Sec. 6.2). In Sec.~\ref{sec:registration_detail} and Sec.~\ref{sec:motion_seg_detail} we explain more details about the experiments for the two applications of scene flow (Sec. 6.3). Lastly in Sec.~\ref{sec:size_and_runtime} we report our model size and runtime and in Sec.~\ref{sec:more_viz} we provide more visualization results on FlyingThings3D and network learned features.

\section{Details on FlyingThings 3D Experiments (Sec. 6.1)}
\label{sec:flying}
The FlyingThings3D dataset only provides RGB images, depth maps and depth change maps. We constructed the point cloud scene flow dataset by popping up 3D points from depth map. The virtual camera intrinsic matrix is 
$$
K = \begin{bmatrix}
f_x=1050.0 & 0.0 & c_x=479.5 \\
0.0 &	f_y=1050.0 & c_y=269.5 \\
0.0 & 0.0 & 1.0
\end{bmatrix}
$$
where $(f_x,f_y)$ are the focal lengths and $(c_x,c_y)$ is the location of principal point. We didn't use RGB images in point cloud experiments.

The $Z$ values of background are significantly larger than the moving objects in the foreground of FlyingThings3D scenes. In order to prevent depth values from explosion and to focus on more apparent motion of foreground objects, we only use points whose $Z$ is larger than a certain threshold $t$. We set $t=35$ in all experiments.

We generate a mask for disappearing/emerging points due to: 1) change of field of view; 2) occlusion. Scene flow loss at the masked points are ignored during training but were used during testing (since we do not have masks at the test time).

\section{Details on Baseline Architectures (Sec. 6.1)}
\label{sec:arch_detail}

\paragraph{FlowNet-C on depth and RGB-D images.}
This model is adapted from \cite{dosovitskiy2015flownet}. The original CNN model takes a pair of RGB images as input. To predict scene flow, we send  a pair of depth images or RGB-D images into the network. Depth maps are transformed to $XYZ$ coordinate maps. RGB-D imags are six-channel maps where the first three channels are RGB images and the rest are $XYZ$ maps. The model has the same architecture as FlowNet-C in \cite{dosovitskiy2015flownet} except that the input has six channels for RGB-D input.

The RGB values are scaled to $[0,1]$. We use the same threshold $t$ as point cloud experiments. Also, scene flow loss at positions where $Z$ value is larger than $t$ are ignored during training and testing. 

\paragraph{EM-baseline.}
The model mixes two point clouds at input level. How to represent the input is not obvious though as two point clouds do not align/correspond.
A possible solution is to append a one-hot vector (with length two) as an extra feature to each point, with $(1,0)$ indicating the point is from the first set and $(0,1)$ for the other set, which is adopted in our EM-baseline.

\begin{figure*}[h]
\centering
\includegraphics[width=0.8\linewidth]{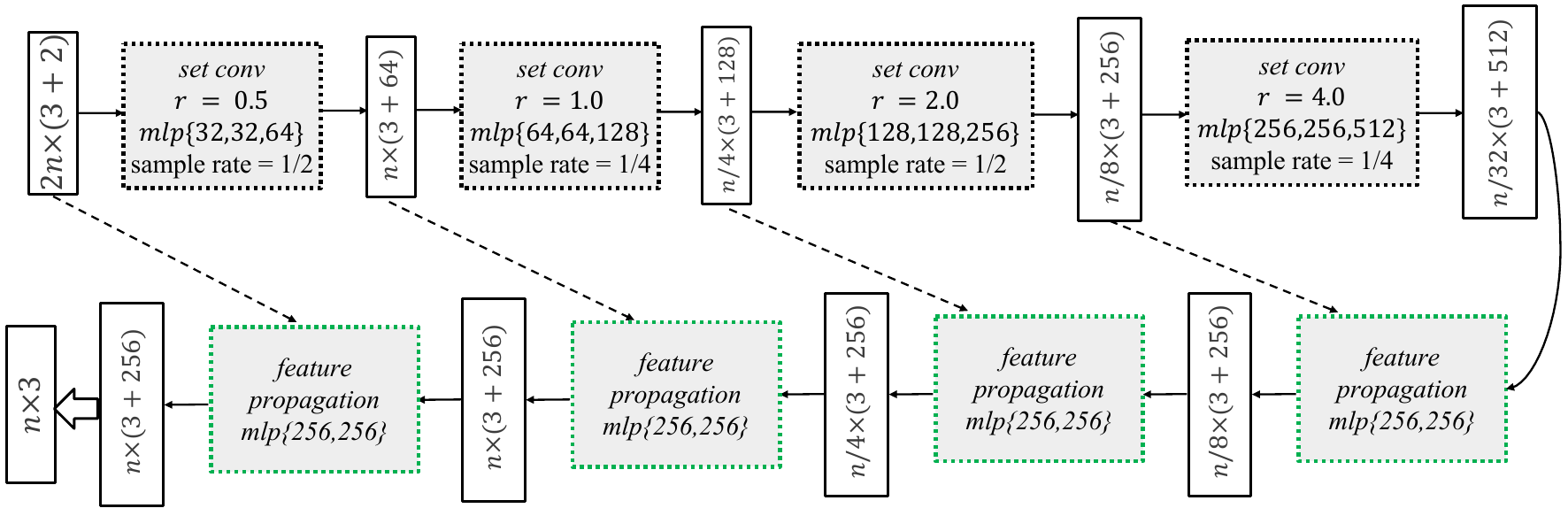}
\caption{Architecture of the Early Mixture baseline model (EM-baseline).}
\label{fig:early_mixture}
\end{figure*}

In Fig.~\ref{fig:early_mixture}, we illustrate our baseline architectures for the EM-baseline.
For each \emph{set conv} layer, $r$ means radius for local neighborhood search, $mlp$ means multi-layer perceptron used for point feature embedding, ``sample rate'' means how much we down-sample the point cloud (for example $1/2$ means we keep half of the original points). The \emph{feature propagation} layer is originally defined in~\cite{qi2017pointnetplusplus}, where features from sub-sampled points are propagated to up-sampled points by 3D interpolation (with inverse distance weights). Specifically, for an up-sampled point its feature is interpolated by three k-NN points in the sub-sampled points. After this step, the interpolated features are then concatenated with the local features linked from the outputs of the set conv layers. For each point, its concatenated feature passes through a few fully connected layers, the widths of which are defined by $mlp\{l_1,l_2,...\}$ in the block.

\begin{figure*}[h]
\centering
\includegraphics[width=0.9\linewidth]{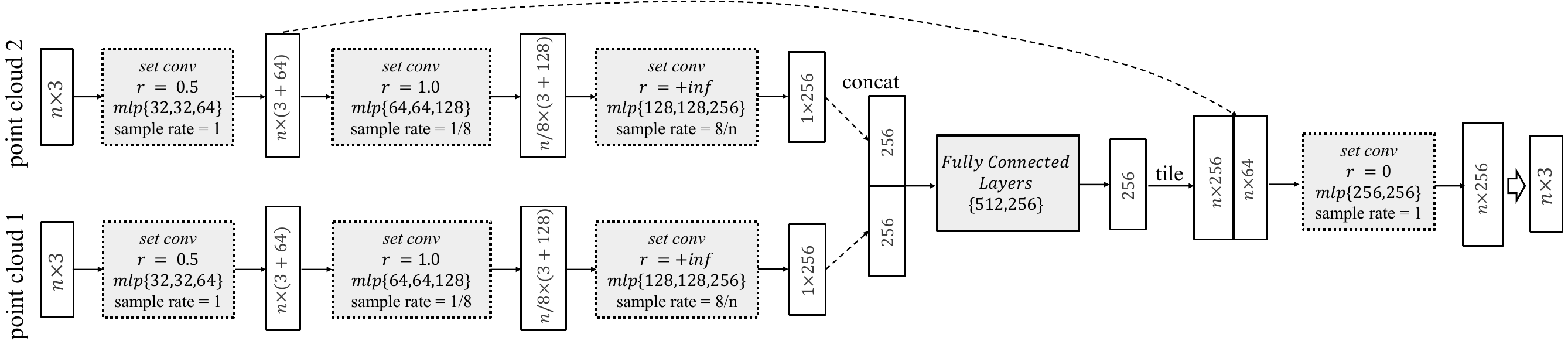}
\caption{Architecture of the Late Mixture baseline model (LM-baseline).}
\label{fig:late_mixture}
\end{figure*}

\paragraph{LM-baseline.}
The late mixture baseline (LM-baseline) mixes two point clouds at the global feature level, which makes it difficult to recover detailed local relations among the point clouds.
In Fig.~\ref{fig:late_mixture}, we illustrate its architecture, which firstly computes global feature from each of the two point clouds, then concatenates the global features and further processes it with a few fully connected layers (mixture happens at global feature level), and finally concatenates the tiled global feature with local point feature from point cloud 1 to predict the scene flow.

\paragraph{DM-baseline.}
While our FlowNet3D model and the DM-baseline both belong to the deep mixture meta architecture, they share the same point feature learning modules to learn intermediate point features and then fix two points at this intermediate level. However they are different in two ways. First the DM-baseline does not adopt a flow embedding layer to ``mix'' the two point clouds (with $XYZ$ coordinates and intermediate features). Instead The DM-baseline concatenates all feature distances and $XYZ$ displacements into a long vector and passes it to a fully connected network before more set conv layers. This however results in sub-optimal learning because it is highly affected by the point orders. Specifically, given a point $p_i = (x_i, f_i)$ in the first point cloud's intermediate point cloud (the one to be mixed with the cloud from the second frame), its $r$ radius neighborhood points in the second frame $\{q_j\}_{j=1}^k$ with $q_j = (y_j, g_j)$, the DM-baseline subsample points in the second frame so that $k$ is fixed and then creates a long vector $v_i \in \mathbb{R}^{2k}$ by concatenation: $(y_j-x_i, d(f_i, g_j))$ for $j=1,...,k$. The function $d$ is a cosine distance function to compute the feature distance of two points. The vector $v_i$ is then processed with a few fully connected layers before feature propagation.
Second, compared to FlowNet3D, the baseline just uses 3D interpolation (with skip links) for flow refinement, with interpolation of three nearest neighborhood with inverse distance weights as described in~\cite{qi2017pointnetplusplus}.

\section{Details on KITTI Data Preparation (Sec. 6.2)}
\label{sec:kitti_prep}
\paragraph{Ground removal.}
For our first evaluation on the KITTI dataset (Table 4 in the main paper), we evaluate on Lidar scans with removed grounds, for two reasons. First, this is a more fair comparison with previous works that relied on ground segmentation/removal as a pre-processing step~\cite{dewan2016rigid,ushani2017learning}. Second, since our model is not trained on the KITTI dataset (due to the very small size of the dataset), it is hard to make it generalize to predicting motions of ground points because the ground is a large flat piece of geometry with little cue to tell its motion. 

\begin{table}
\centering
\scalebox{1.0}{
\begin{tabular}{ c|c|c }
\toprule
Method  & RANSAC & GroundSegNet   \\ \midrule
Accuracy & 94.02\%  & 97.60\%  \\ \midrule
Time per frame & 43 ms& 57 ms\\ 
\bottomrule
\end{tabular}}
\caption{Evaluation for ground segmentation on KITTI Lidar scans.  Accuracy is averaged across test frames.}
\label{sceneflow:tab:ground_removal}
\end{table}

To validate we can effectively remove grounds in LiDAR point clouds, we evaluate two ground segmentation algorithms: RANSAC and GroundSegNet. RANSAC fits a tilted plane to point clouds and classify points close to the plane as ground points. GroundSegNet is a PointNet segmentation network trained to classify points (in 3D patches) to ground or non-ground (we annotated ground points in all 150 frames and used 100 frames as train and the rest as test set).
Both methods can run in real time: 43ms and 57ms per frame respectively, and achieve very high accuracy: \emph{94.02\%} and \emph{97.60\%} averaged across test set.
Note that for evaluation in the main paper Table 4, we used our annotated ground points for ground removal, to avoid dependency on the specific ground removal algorithm.

\paragraph{Inference on large point clouds.}
On large KITTI scenes, we split the scene into multiple chunks. Chunk positions are the same for both frames. Each chunk has size of 5m$\times$5m and is aligned with XY axes (considering Z is the up-axis). There are overlaps between chunks. In practice, neighboring chunks are off by 2.5m with a small noise (Gaussian with 0.3 std) in $X$ or $Y$ direction to each other. 

We run the final FlowNet3D model on pairs of frame 1 chunk and frame 2 chunk that are at the same location. 
Points appearing in more than one chunk have their estimated flows averaged to get the final output. 

\section{Details on the Scan Registration Application (Sec. 6.3.1)}
\label{sec:registration_detail}
For this experiment we prepared a partial scan dataset by virtually scanning the ModelNet40~\cite{wu20153d} CAD models with a rotated camera around the center axis of the object, with the same train/test split as for the classification task. The virtual scan tool is provided by the Point Cloud Library. In partial scans, parts of an object may appear in one scan but missing in the other, which makes registration/warping very challenging. 

We finetuned our FlowNet3D model on this dataset, to predict the 3D warping flow from points in one partial scan to their expected positions in the second scan. Then at inference time, we predict the flow for each point in the first scan as its scene flow. Since the point moving distance can be very large in those partial scans, we iteratively regress twice for the scene flow (i.e.  predict a flow from point cloud 1 to point cloud 2, and then predict a second residual flow from point cloud 1 + first flow to point cloud 2). Then the final scene flow is the 1st flow + the residual flow (visualized in Fig. 6 main paper). To get a rigid motion estimation from the scene flow, we can fit a rigid transformation from the point cloud 1 to the point cloud 2 + scene flow, as they have one-to-one correspondences. Then the rigidly transformed point cloud 1 is the final estimation of our warping (shown in main paper Fig. 6 right while the warping error is reported in main paper Table 6).

\begin{figure*}[h]
\centering
\includegraphics[width=0.7\linewidth]{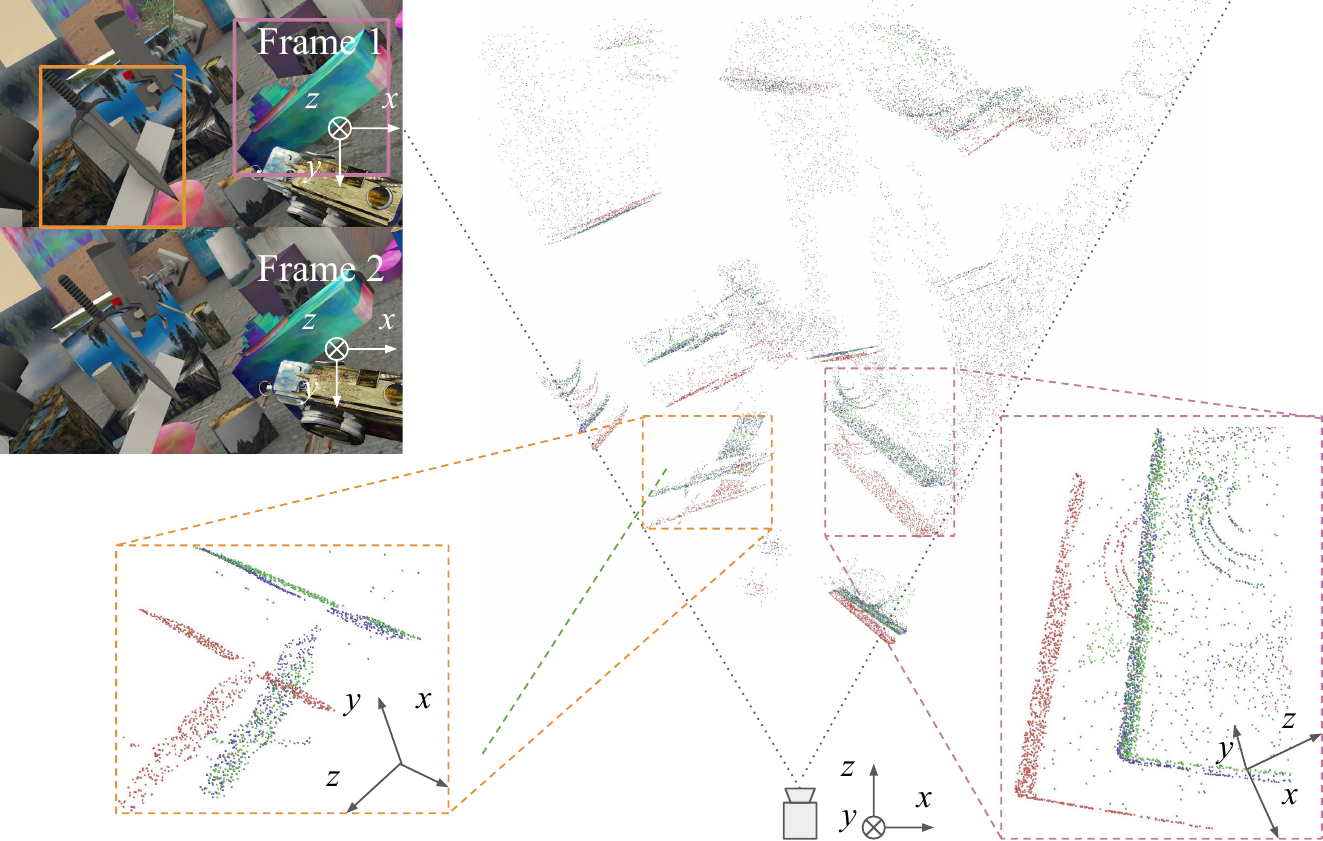}
\caption{Scene flow results for TEST-A-0061-right-0013  of FlyingThings3D.}
\label{fig:flying:viz:1}
\end{figure*}

\begin{figure*}[h]
\centering
\includegraphics[width=0.7\linewidth]{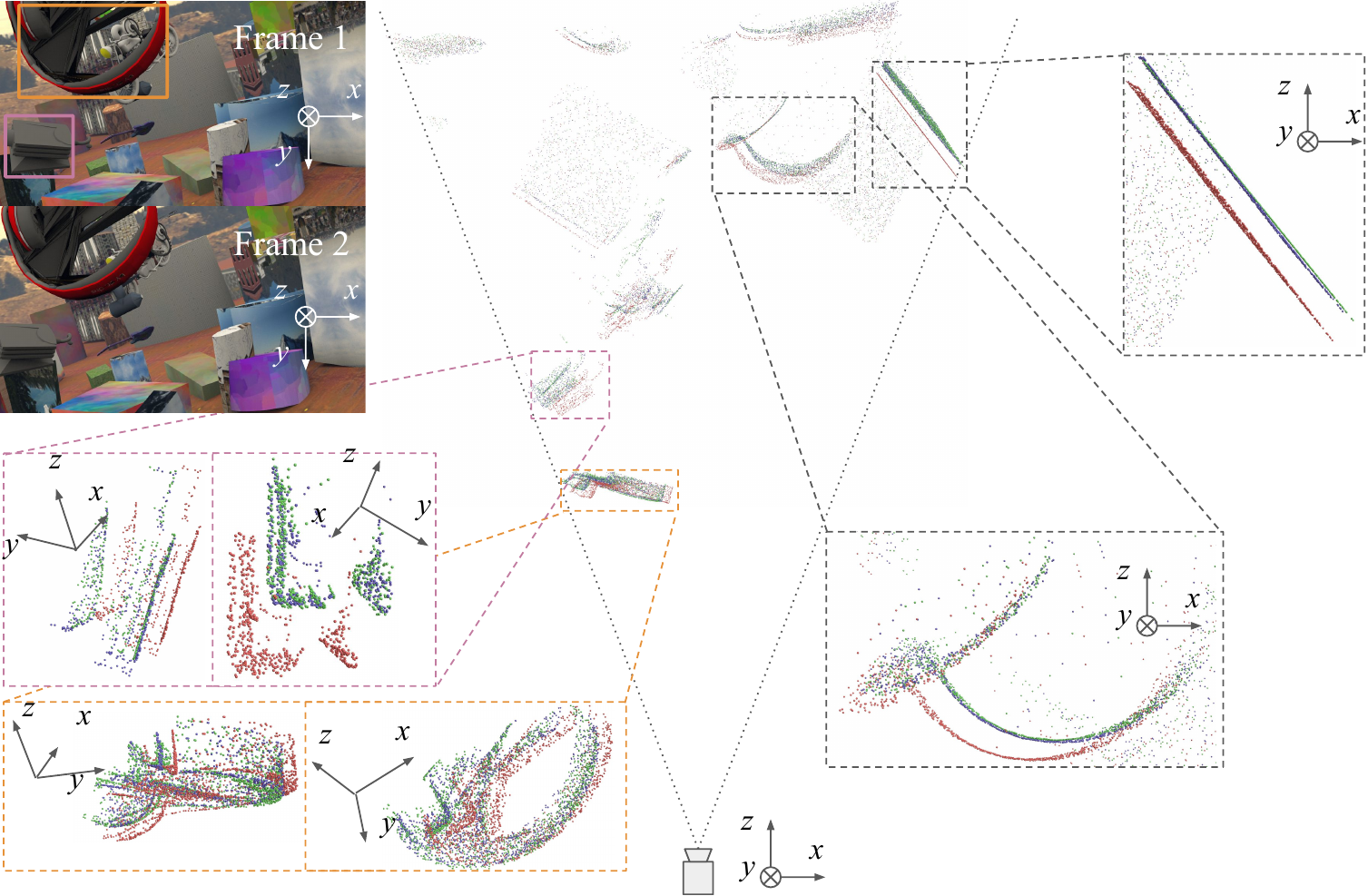}
\caption{Scene flow results for TEST-A-0006-right-0011 of FlyingThings3D.}
\label{fig:flying:viz:2}
\end{figure*}

\begin{figure*}[h]
\centering
\includegraphics[width=0.7\linewidth]{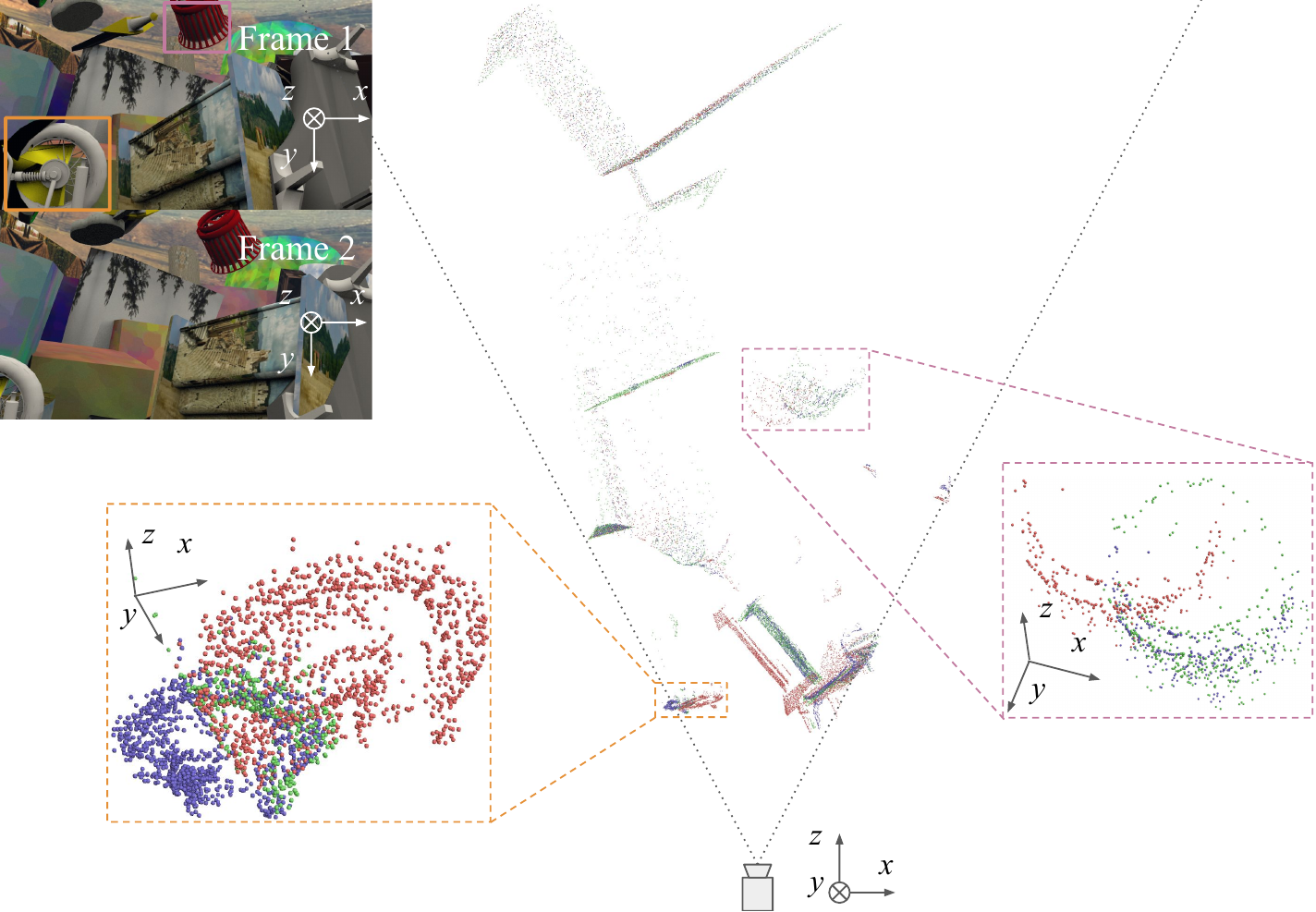}
\caption{Scene flow results for TEST-B-0011-left-0011 of FlyingThings3D.}
\label{fig:flying:viz:3}
\end{figure*}

\section{Details on the Motion Segmentation Application (Sec. 6.3.2)}
\label{sec:motion_seg_detail}

We first obtained the estimated scene flow with the method discussed in Sec. \ref{sec:kitti_prep}. Then the flow is multiplied with a factor $\lambda$ and is concatenated with coordinates of each point as a 6-dim vector $(x,y,z,\lambda d_x,\lambda d_y,\lambda d_z)$. Next based them we find connected components in the 6-dim space by setting two hyperparamters: a proper minimum cluster size and distance upper bound for forming a cluster.

\section{Model Size and Runtime}
\label{sec:size_and_runtime}
FlowNet3D has a model size of 15MB, which is much smaller than most deep convolutional neural networks. In Table~\ref{tab:speed}, we show the inference speed of the model on point clouds with different scales. For this evaluation we assume both point clouds from the two frames have the same number of points as specified by \#points. We test the runtime on a single NIVIDA 1080 GPU with TensorFlow~\cite{abadi2016tensorflow}.

\begin{table}[h!]
\small
\centering
\resizebox{\linewidth}{!}{
\begin{tabular}{c|ccccccc}
\toprule
\#Points & 1K & 1K & 2K & 2K & 4K & 4K & 8K \\ \midrule
Batch size & 1 & 8 & 1 & 4 & 1 & 2 & 1 \\ \midrule
Time (ms) &  18.5  & 43.7 & 36.6 & 58.8 &  101.7   &  117.7 & 325.9       \\ \bottomrule
\end{tabular}}
\caption{Runtime of FlowNet3D with different input point cloud sizes and batch sizes. For this evaluation we assume the two input point clouds have the same number of points.}
\label{tab:speed}
\end{table}

\begin{figure*}[t]
\centering
\includegraphics[width=0.95\linewidth]{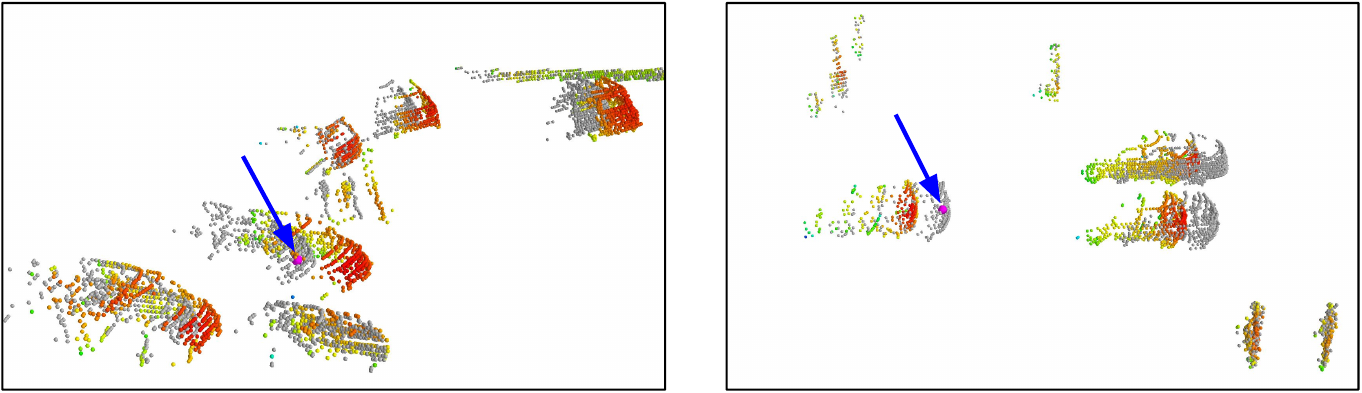}
\caption{Visualization of local point feature similarity. Given a \textcolor{magenta}{point $P$} (pointed by the blue arrow) in frame 1 (gray), we compute a heat map indicating how points in frame 2 are similar to $P$ in feature space. More red is more similar.}
\label{fig:feature_similarity}
\end{figure*}

\begin{figure*}[t]
    \centering    
    \includegraphics[width=0.99\textwidth]{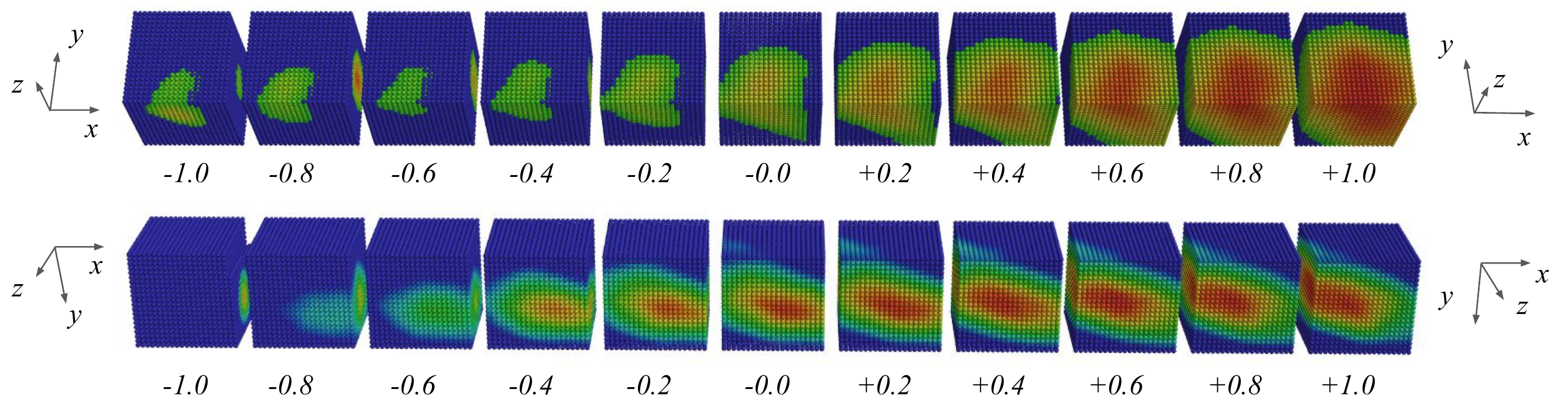}   
        \caption{Visualization of flow embedding layer. Given a certain similarity score (defined by one minus cosine distance, at the bottom of each cube), the visualization shows which $(x,y,z)$ displacement vectors in a $[-5,5]\times [-5,5] \times [-5,5]$ cube activate one output neuron of the flow embedding layer.}
\label{fig:flow_embedding_viz}
\end{figure*}

\section{More Visualizations}
\label{sec:more_viz}

\paragraph{Visualizing scene flow results on FlyingThings3D}
We provide results and visualization of our method on FlyingThings3D test set \cite{Flyingthings3D:Driving}. The dataset consists of rendered scenes with multiple randomly moving objects sampled from ShapeNet \cite{ShapeNet}. To clearly visualize the complex scenes, we provide the view of the whole scene from top. We also  zoom in and view each object from one or more directions. The directions can be inferred from  consistent $XYZ$ coordinates shown in both the images and point cloud scene. We show points from \textcolor{red}{frame 1}, \textcolor{green}{frame 2} and \textcolor{blue}{estimated flowed points} in different colors.  Note that local regions are zoomed in and rotated for clear viewing. To help find  correspondence between images and point clouds, we used distinct colors for zoom-in boxes of corresponding objects. Ideal prediction would roughly align blue and green points. The results are illustrated in Figure \ref{fig:flying:viz:1}-\ref{fig:flying:viz:3}.

Our method can handle challenging cases well. For example, in the orange zoom-in box of Figure \ref{fig:flying:viz:1}, the gray box is occluded by the sword in both frames and our network can still estimate the motion of both the sword and visible part of the gray box well. There are also failure cases, mainly due to the change of visibility across frames. For example, in the orange zoom-in box of Figure \ref{fig:flying:viz:3}, the majority of the wheel is visible in the first frame but not visible in the second frame. Thus our network is confused and the estimation of the motion for the non-visible part is not accurate.

\paragraph{Network visualization}

Fig.~\ref{fig:feature_similarity} visualizes the local point features our network has learned, by showing a heatmap of correlations between a chosen point in frame 1 and all points in frame 2. We can clearly see that the network has learned geometric similarity and is robust to partiality of the scan.

Fig.~\ref{fig:flow_embedding_viz} shows what has been learned in a flow embedding layer. Looking at one neuron in the flow embedding layer, we are curious to know how point feature similarity and point displacement affect its activation value. To simplify the study, we use a model trained with cosine distance function instead of network learned distance (through directly inputing two point feature vectors). We iterate distance values and displacement vector, and show in Fig.~\ref{fig:flow_embedding_viz} that as similarity grows from -1 to 1, the activation becomes significantly larger. We can also see that this dimension is probably responsible for a flow along the positive $Z$ direction.

\newpage

\end{document}